\documentclass{article}

\usepackage{arxiv}

\usepackage[utf8]{inputenc} % allow utf-8 input
\usepackage[T1]{fontenc}    % use 8-bit T1 fonts
\usepackage{hyperref}       % hyperlinks
\usepackage{url}            % simple URL typesetting
\usepackage{booktabs}       % professional-quality tables
\usepackage{amsfonts}       % blackboard math symbols
\usepackage{nicefrac}       % compact symbols for 1/2, etc.
\usepackage{microtype}      % microtypography
\usepackage{lipsum}

\usepackage{multirow} % by DG
\usepackage{pifont} % by DG
\usepackage{graphicx} 
\usepackage{hyperref}
\title{Imponderous Net for Facial Expression Recognition in the Wild}

\author{
  Darshan Gera \\
  SSSIHL, DMACS, Brindavan, 560067, India\\
  \texttt{darshangera@sssihl.edu.in} \\
   \And
S Balasubramanian \\
  SSSIHL, DMACS, Puttaparthi, 515134, India \\
  \texttt{sbalasubramanian@sssihl.edu.in} \\
}

\begin{document}
\maketitle
%%%%%%%%%%%%%%%%%%%%%%%%%%%%%%%%%%%%%%%%%%%%%%%%%%%%%%%%%%%%%%%%%%%%%%%%%%%%%%%%
\begin{abstract}

Since the renaissance of deep learning (DL), facial expression recognition (FER) has received a lot of interest, with continual improvement in the performance. Hand-in-hand with performance, new challenges have come up. Modern FER systems deal with face images captured under uncontrolled conditions (also called in-the-wild scenario) including occlusions and pose variations. They successfully handle such conditions using deep networks that come with various components like transfer learning, attention mechanism and local-global context extractor. However, these deep networks are highly complex with large number of parameters, making them unfit to be deployed in real scenarios. Is it possible to build a light-weight network that can still show significantly good performance on FER under in-the-wild scenario? In this work, we methodically build such a network and call it as Imponderous Net. We leverage on the aforementioned components of deep networks for FER, and analyse, carefully choose and fit them to arrive at Imponderous Net. Our Imponderous Net is a low calorie net with only 1.45M parameters, which is almost 50x less than that of a state-of-the-art (SOTA) architecture. Further, during inference, it can process at the real time rate of 40 frames per second (fps) in an intel-i7 cpu. Though it is low calorie, it is still power packed in its performance, overpowering other light-weight architectures and even few high capacity architectures. Specifically, Imponderous Net reports 87.09\%, 88.17\% and 62.06\% accuracies on in-the-wild datasets RAFDB, FERPlus and AffectNet respectively. It also exhibits superior robustness under occlusions and pose variations in comparison to other light-weight architectures from the literature.

\end{abstract}

%%%%%%%%%%%%%%%%%%%%%%%%%%%%%%%%%%%%%%%%%%%%%%%%%%%%%%%%%%%%%%%%%%%%%%%%%%%%%%%%
\section{INTRODUCTION}

Recognizing expressions in faces plays a vital role in communication and social interaction, analysing mental related illness like depression, measuring attentiveness in student-teacher interaction etc. Traditional works like \cite{56,60} focused on training machines for FER through examples collected in a controlled (in-lab) environment. Examples of such in-lab datasets are CK+ \cite{63_a,63_b} , Oulu-CASIA \cite{61} and JAFFE \cite{58}. Due to the resurgence of deep neural networks (DNNs), a significant improvement has been achieved in FER systems under controlled environment \cite{40, chen2019facial_51}. 

DNNs also handle well new challenges in the uncontrolled environment including occlusions and pose variations \cite{scan,oadn, ran, gacnn}. However, these networks are deep, imbedded with a large number of parameters (for e.g. 70M parameters in \cite{scan}). Such networks are unfit to be deployed in real scenarios. For e.g., in a driver alert system, recognizing the drowsy state of driver and raising alert in real time is crucial to prevent accidents. Another example is in the deployment of FER system in the wearable tool to assist autistic children to understand social situations in real-time \cite{autistic}. 

In this work, we build methodically a low calorie, power packed network to perform FER under in-the-wild scenario efficiently. We call this network Imponderous Net. We identify the important components in SOTA deep architectures for in-the-wild FER, analyse them, carefully choose and fit them to arrive at Imponderous Net. Imponderous Net has only 1.45M parameters, which is almost 50x less than that of SOTA SCAN architecture \cite{scan}. Further, during inference, it can process at the real time rate of 40 frames per second (fps) in an intel-i7 cpu. Though it is low calorie, it is still power packed in its performance, overpowering other light architectures and even few high capacity architectures. Specifically, Imponderous Net reports 87.09\%, 88.17\% and 62.06\% accuracies on in-the-wild datasets RAFDB \cite{rafdba, rafdbb}, FERPlus \cite{ferplus} and AffectNet \cite{affectnet} respectively. It also exhibits superior robustness under occlusions and pose variations in comparison to other light-weight architectures from the literature. In summary, our contributions are:

\begin{enumerate}
    \item Methodically analyse the modern deep architectures for in-the-wild FER, identify important components in them, carefully choose and fit them to arrive at Imponderous Net, a low calorie power packed architecture for in-the-wild FER.
    \item Exhibit the performance of Imponderous Net against a variety of light and heavy DNNs for FER.
    \item Highlight the performance of Imponderous Net under challenging conditions like occlusions and pose variations. 
    \item We do not shy away from reporting comparison of our Imponderous Net against current relatively heavy SOTA architectures for FER under in-the wild setting, even though in some places, some of them are relatively far ahead of the light-weight architectures. We do this to clearly point out that certain challenging conditions demand more parameters. We believe this kind of reporting will be very useful to the research community.
\end{enumerate}

\section{Related work}
\label{rel_work}
We focus on FER under in-the-wild scenario. In \cite{gacnn}, unobstructedness or importance scores of \textbf{local patches} of feature maps corresponding to certain landmark points are computed using \textbf{self-attention}, and the respective local feature maps are weighted by these scores. The expectation is that, over training, patches corresponding to occluded areas in the image will receive low importance and hence become less relevant. Parallely, \textbf{global context} is captured through self attention on the entire feature map. Concatenation of \textbf{local-global context} is passed to a classifier for expression recognition. Region attention network (RAN) \cite{ran} is conceptually similar to \cite{gacnn} but selects patches directly from the image input. RAN combined with a region biased loss quantifies the importance of patches. Subsequently, a \textbf{relation-attention} module that relates local
and global context provides the expression representation for classification.  In \cite{oadn}, \textbf{attention} weights are generated as samples from a Gaussian distribution centered at spatial locations of the feature map, corresponding to certain confident landmark points, where the confidence score is provided by an external landmark detector. Selection of local patches follow \cite{gacnn}. Concurrently, complementary information is gathered from non-overlapping partitions of the feature map. Together, the patch based information and the complementary information guide the classifier to report state-of-the-art results. Unlike \cite{oadn, ran} and \cite{gacnn}, \cite{scan} uses a \textbf{local-global attention} branch that computes attention weight for every channel and every spatial location across certain local patches and the whole input to make FER model robust to occlusions and pose variations. It does not require external landmark detector. It is to be noted that all the SOTA methods \cite{scan,oadn, ran, gacnn} have a \textbf{base model that is pre-trained on FR} and a subsequent \textbf{attention mechanism}.

Though the current methods \cite{scan, oadn, ran, gacnn} have enhanced the performance under challenging conditions like occlusions and pose variations, they come with the heavy baggage of large number of parameters. There have been some efforts recently to make the FER models light-weight \cite{microexpnet, facechannel, emotionnanonet}, but a majority of them \cite{microexpnet, emotionnanonet} report performance on in-lab datasets only, and some of them \cite{emotionnanonet} on only one dataset. MicroExpNet \cite{microexpnet} distills a tiny student model from an inception-v3 model \cite{microexpnet} for FER on two in-lab datasets. Facechannel \cite{facechannel} deploys a light-weight CNN that has an inhibitory layer connected to the last layer of the network to help shape learning of the facial features. EmotionNet Nano \cite{emotionnanonet} is created using a two-phase design strategy. In the first phase, residual architecture design principles are leveraged to capture the complex nuances of facial expressions. In the second phase, machine-driven design exploration is employed to generate the final tailor-made architecture. The method is tested on only one in-lab dataset.

\section{Building the Imponderous Net}
It can be observed from section \ref{rel_work} that the success of the recent SOTA methods \cite{scan,oadn, ran, gacnn} for FER under in-the-wild scenario largely depends on three important components viz. (i) transfer learning (finetuning) from FR domain, (ii) some kind of attention mechanism to facilitate the model focus on relevant regions for FER, and (iii) using information from both local and global context to decipher discriminative features. 
This observation led us to ensure that the Imponderous Net is built on the foundations of the three aforementioned building blocks, though in its own way, as will be discussed further. Apart from this, Imponderous Net also has implicit ensembling to up the ante. We will now discuss the building blocks of Imponderous Net.

\begin{table*}
\caption{Base model summary in recent SOTA methods}
\label{Table 1}
\begin{center}
\begin{tabular}{|c||c|c|c|}
\hline
Method & Base model & Pre-trained on & No. of params\\
\hline
gACNN \cite{gacnn} & VGG-16 (up to 9th conv layer) & ImageNet \cite{imagenet_cvpr09_50} & 5.27M\\
RAN \cite{ran} & ResNet-18 (up to last pooling layer) & ms-celeb-1M \cite{17} & 11.18M\\
OADN \cite{oadn} & ResNet-50 (up to 4th conv block) & VGGFace2 \cite{20} & 23.5M\\
SCAN \cite{scan} & ResNet-50 (up to 3rd conv block) & VGGFace2 & 8.5M\\
\hline
\end{tabular}
\end{center}
\end{table*}

\subsection{Building block I - Transfer learning}
To choose the base model for Imponderous Net, we first looked at the base models of the recent SOTA methods \cite{scan,oadn, ran, gacnn}. This is summarized in Table \ref{Table 1}. The base models themselves have relatively large number of parameters, adding significant amount of fat to the whole model. Our first endeavor is to have a light-weight base model. Though transfer learning plays a crucial role, the amount of knowledge transferred and its relevance impacts the performance. Particularly, for FER, identity specific features from FR are not relevant. What is relevant is the facial features \cite{scan}. Such features are generally available in the middle layers of DNN \cite{cnn_features}. So, unlike \cite{ran} and \cite{oadn} that extract features from the last convolutional/pooling layers of the base model, we can focus on the middle layers. In fact, \cite{gacnn} and \cite{scan} follow this idea. However, in \cite{gacnn}, VGG-16 by itself is a heavy model. \cite{scan} also has relatively a large number of parameters, totalling to 8.5M, in the base model. 

\begin{figure}
\centerline{\includegraphics[width=2in]{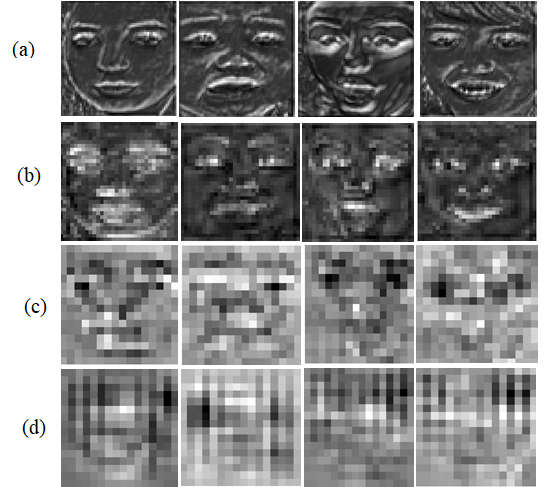}}
\caption{Feature maps from the four levels of lightcnn29 architecture. Each row corresponds to one level, in the ascending order.}
\label{lightcnn_fmaps}
\end{figure}

Towards identifying a lighter base model, we choose LightCNN \cite{lightcnn} designed for FR. LightCNN has proved its success as a light-weight model for FR. The best version of it (lightcnn29) has around 12M parameters. However, we require access upto only middle layers. We choose as facial features the output feature maps from the pooling layer following MFM3 in the lightcnn29 architecture \cite{lightcnn}. MFM stands for max-feature-map, a non-linearity introduced in LightCNN, whose influence is discussed in the next paragraph. The overall size of base model in the Imponderous Net amounts to only 1.18M parameters. To validate our selection policy, we plot the feature maps from the four levels of layers in lightcnn29 architecture in Fig. \ref{lightcnn_fmaps}. It is clear that, while the lower level (first row in the figure) captures low level features like edges, the middle level layers (second row in the figure) capture parts like eyes, mouth, nose tip etc. As we go higher (3rd and 4th rows in the figure), the resolution reduces, and the whole face is captured. We choose middle level features which correspond to the pooling layer following MFM3 in the lightcnn29 architecture \cite{lightcnn}.

It is known from neural science that lateral inhibition \cite{lightcnn_ihbition} increases the contrast and sharpness in visual response. What this means is that if there is a vertical edge in an image, the neuron excited by the vertical edge sends an inhibitory signal to all its neighboring neurons, thereby increasing its visual response to the vertical edge. MFM in LightCNN aims to mimick this feature. This behaviour helps to separate the informative signals from the noisy signals. In fact, MFM has a significant role in cleaning the ms-celeb-1M FR dataset \cite{lightcnn}.  It is very important that we have the pre-trained model on a dataset with clean labels because large datasets generally come with noisy labels, and it is well known that DNNs tends to memorize noisy labels \cite{dn_memorize}. It is to be noted that one of the light-weight models for FER \cite{facechannel} argues about the importance of inhibitions and introduces an associated inhibitory layer with the last convolutional layer in its model through an extra set of convolutions. However, in our base model, we get it free with LightCNN architecture, without adding any extra set of convolutions.   In summary, the base model of the Imponderous Net is shown in the Table \ref{Table 2}.

\begin{table}
\caption{Base model in the Imponderous Net}
\label{Table 2}
\begin{center}
\begin{tabular}{|c||c|c|}
\hline
Type & Filter size & o/p size \\
\hline
Conv1 & 5 x 5/1, 2 & 128 x 128 x 96\\
MFM1  & - & 128 x 128 x 48\\
\hline
Pool1 & 2 x 2/2 & 64 x 64 x 48 \\
\hline
Conv2\_x  &$\left [ \begin{array}{cc} 3 \times 3/1, & 1  \\ 3 \times 3/1, & 1 \end{array} \right] \times 1$ & 64 x 64 x 48 \\
Conv2a & 1 x 1/1 & 64 x 64 x 96 \\
MFM2a & - & 64 x 64 x 48 \\
Conv2  & 3 x 3/1, 1 & 64 x 64 x 192 \\
MFM2 & - & 64 x 64 x 96 \\
\hline
Pool2 & 2 x 2/2 & 32 x 32 x 96\\
\hline
Conv3\_x  &$\left [ \begin{array}{cc} 3 \times 3/1, & 1  \\ 3 \times 3/1, & 1 \end{array} \right] \times 2$ & 32 x 32 x 96 \\
Conv3a & 1 x 1/1 & 32 x 32 x 192 \\
MFM3a & - & 32 x 32 x 96 \\
Conv3  & 3 x 3/1, 1 & 32 x 32 x 384 \\
MFM3 & - & 32 x 32 x 192 \\
\hline
Pool3 & 2 x 2/2 & 16 x 16 x 192\\
\hline
\end{tabular}
\end{center}
\end{table}

\begin{figure*}
\centerline{\includegraphics[width=5in]{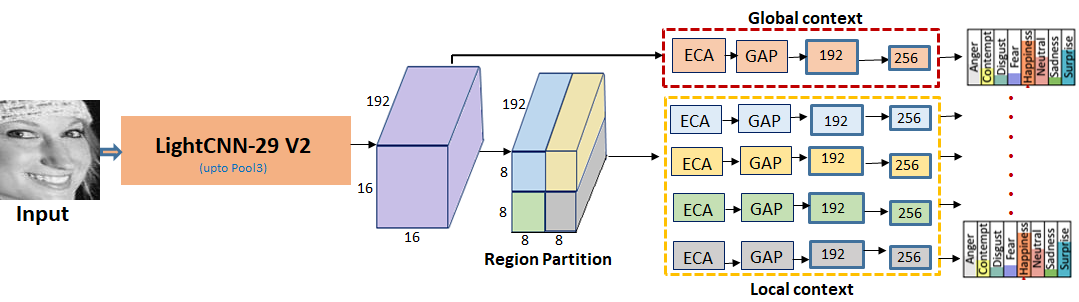}}
\caption{The Imponderous Net Architecture. Note that GAP stands for global average pooling.}
\label{framework}
\end{figure*}
\subsection{Building block II - Attention Mechanism}
Attention mechanism has become a key component \cite{scan,oadn, ran, gacnn} for FER under in-the-wild scenario to avoid occlusions, and handle pose variations. It is implemented using only dense layers or convolutional layers or a combination of both. The input to the attention unit is constructed from the output of the base model. The attention weight is either constant across both spatial and channel dimensions \cite{gacnn}, or constant across spatial dimensions \cite{ran}, or constant across channel dimensions \cite{oadn}, or specific to each spatial and channel dimension \cite{scan}. 

Our goal is to choose an attention mechanism that is efficient (facilitates handling occlusions and pose variations), and at the same time does not bloat up the dainty base model we had arrived at earlier. Towards this end, we look at the capacity of the attention units in the recent SOTA methods \cite{scan,oadn, ran} and \cite{gacnn}. Table \ref{Table 3} shows the numbers. The whopping rise in \cite{gacnn} and \cite{scan} in comparison to the negligible additions in \cite{oadn} and \cite{ran} is because of two reasons: (i) convolutional based attention mechanism at multiple local patches (around 25 of them) in both \cite{gacnn} and \cite{scan}, (ii) specific attention weight to each channel and spatial unit in \cite{scan}. In fact, \cite{scan}, which gives current SOTA results,  reports that around 2\% loss will be incurred in the performance if attention is not incorporated at the level of multiple local patches. However, given the goal of building a light-weight model for FER under in-the-wild scenario, we cannot bloat up the attention unit for a small push up in the performance. We could use attention units from \cite{ran,oadn}. But \cite{ran} requires multiple crops of the input, and \cite{oadn} requires external landmark detector. Both these requirements would scale the number of floating point computations. 

Instead, we rely on a very light-weight attention mechanism called efficient channel attention (ECA) \cite{eca}, which is specifically designed to overcome the paradox of performance and complexity trade-off mentioned above. To obtain attention weights per channel, cross-channel interaction is very important. Most attention units \cite{gacnn} follow \cite{sqe} wherein, to lower complexity, dimensionality reduction is performed while incorporating cross-channel interaction. ECA \cite{eca} shows that lowering dimensionality in attention unit has negative impact on performance. In fact, ECA avoids dimensionality reduction by using a simple  1-d convolution with an adaptive filter size for incorporating cross-channel interaction. This hardly adds any parameters. In our Imponderous Net, ECA adds an extra 4000 parameters only. 

\begin{table}
\caption{Attention unit summary in recent SOTA methods}
\label{Table 3}
\begin{center}
\begin{tabular}{|c||c|}
\hline
Method & No. of params\\
\hline
gACNN \cite{gacnn} &  $>$ 100M\\
RAN \cite{ran} & Negligible\\
OADN \cite{oadn} & Around 1M\\
SCAN \cite{scan} & Around 60M\\
\hline
\end{tabular}
\end{center}
\end{table}

\subsection{Building block III - Local and global context}
Processing the output from the base model as a whole alone is not enough for FER under in-the-wild scenario since this would not avoid information from the occluded regions. This is the primary reason why all the SOTA methods \cite{gacnn, ran, oadn, scan} employ attention mechanism at multiple local patches. While some \cite{gacnn, scan} do it in a sophisticated manner, others \cite{ran, oadn} rely on multiple input crops or external landmark detectors to define it in a simple manner. In our Imponderous Net, neither do we crop the input multiple times nor we depend on external landmark detector for local patch processing. Instead, we take cue from the complementary context information (CCI) branch of \cite{scan} wherein the output from base model is partitioned into equal sized non-overlapping blocks. For us, these non-overlapping blocks provide the local context. The sufficiency of this selection is supported by an experiment in \cite{scan} wherein the authors have shown that CCI branch carries larger weightage with regard to the overall performance. We have four 8 x 8 x 192 dimensional local patches since the output from the base model is of size 16 x 16 x 192. We employ ECA to each of these blocks. 

Along with local context, we also consider the whole output feature map from the base model. This provides the global context. ECA is applied to the global context as well. To enhance the discriminatory power of facial expression features, feature diversity is important. With both local and global context being processed concurrently, this is likely to be achieved.

\subsection{Other considerations}
Now that we have the building blocks ready, we complete the architecture by incorporating a dense layer that gives the facial expression features, and a subsequent classification layer for expression recognition. The complete architecture is shown in Fig. \ref{framework}. An important point to note is that each patch (local or global) is supervised separately. This implicitly provides an ensemble of supervisions which has the inherent potential to act as a regularizer and boost the performance. The implicit ensembling does not bloat the model unlike explicit ensembling \cite{Hassner1} wherein the entire network has to be replicated. 

%We also employ two other simple tricks to enhance the performance. The first is data augumentation during training. Instead of manually specifying augumentations like random horizontal flip etc., we use auto-augument \cite{autoaugument}, a mechanism for automatically finding the appropriate data augumentation policies for the given data. We found it to be very effective. The second trick we use is the mirror trick wherein we pass a couple of images as input - the image and its horizontal flip. We concatenate the features from both of them before classification.

\section{Datasets and implementation details}
\subsection{Datasets}
The in-the-wild datasets considered are AffectNet \cite{affectnet}, RAFDB \cite{rafdba, rafdbb}, FERPlus \cite{ferplus} and FED-RO \cite{gacnn}.
AffectNet is the largest facial expression dataset with 1M images out of which 0.44M are manually annotated and remaining 0.46M images are automatically annotated for the presence of eight (neutral, happy, angry, sad, fear, surprise, disgust, contempt) facial expressions. We do not consider the contempt expression in this dataset.

RAFDB contains 29762 facial images tagged with basic or compound expressions by 40 annotators. In this work, we use the subset with 7-basic emotions consisting of 12,271 images for training and 3068 images for testing.

FERPlus, an extended version of FER2013, consists of 28709 images for training, 3589 images for validation and 3589 for testing with all 8-basic emotions.  

We also evaluate the performance on the challenging subsets of AffectNet, RAFDB and FERPlus with regard to occlusions and pose variations greater than 30 and 45 degrees \cite{ran}. For details on the statistics of these subsets, readers are referred to \cite{scan}. We also present the results on the real occlusions dataset, FED-RO \cite{gacnn}.  %Number of images in these subsets are presented in Table \ref{tab:Table 4}. 
\iffalse
 \begin{table}
    \centering 
    \caption{Number of images in challenging test subsets} 
    
    \begin{tabular}{|c|c|c|c|}
         \hline
          Dataset      & Occlusion   & Pose $>$ 30 & Pose $>$ 45  \\
         \hline
         \hline
         FERPlus     &  605         & 1171 & 634 \\
         RAFDB       &  735         & 1248 & 558 \\
         AffectNet   &  592          & 1614 & 822 \\
         FEDRO       &  400          &   -   &   -  \\    
         \hline 
    \end{tabular}
    \label{tab:Table 4}
\end{table}
\fi
\subsection{Implementation details}
Implementation is done in Pytorch. Face images are detected and aligned using MTCNN \cite{mtcnn}. These are further converted to grayscale and resized to 128 x 128. Conversion to grayscale promotes invariance to illumination variations \cite{lightcnn}. Our base network has been pre-trained on clean ms-celeb-1M \cite{17} and casia-webface \cite{casia} datasets. It is further finetuned during training.  Batch size is set to 64. The whole network is trained using Adamax optimizer. Learning rate (lr) for base network is 0.001. For the rest of the network, it is assigned 0.01. Weight decay is fixed at 4e-5. Data augmentation is done using an automatic policy \cite{autoaugument}. Oversampling is adopted to overcome imbalance problem on AffectNet dataset. Mirror trick (i.e both the image and its horizontal flip are considered as inputs) is employed during both training and testing. Evaluation metric used is overall accuracy across all the datasets. Note that, for fair comparison against other light-weight methods, we employed mirror trick to them as well. With regard to data augmentation, we followed their work. In the case where no data augmentation is used in their work, we report the best result among 'no augmentation' or 'auto-augmentation'.

\section{Results and discussions}
\subsection{Performance comparison with SOTA methods for FER under in-the-wild setting}
Table \ref{Table 5} presents the comparison. Our Imponderous Net has outperformed the high capacity gACNN by a significant margin of around 2.5 to 3.5\% on all the datasets. It has also overtaken FMPN by a large margin. Further, it has outdone RAN in RAFDB and AffectNet datasets. It also performs better than SCN \cite{scn} on FERPlus dataset. It trails the current SOTA method SCAN by only 1.2\% and 2\% on FERPlus and RAFDB, respectively. On AffectNet, the Imponderous Net, though lags behind SCAN by 3\%, it still has breached 60\% mark, which only a few methods in the literature has done currently. Given that Imponderous Net has only 1.45M parameters, which is almost 50 times lesser then the number of parameters SCAN has, its performance is definitely power packed. Note that AffectNet is the largest FER dataset, and hence it is likely to contain noisy labels due to its sheer size. In fact, SCN \cite{scn}, though has relatively lesser number of parameters than SCAN, has done exceedingly well on AffectNet because it explicitly handles noisy labels and corrects them. We believe that the performance of our Imponderous Net in AffectNet can be raised provided it can handle noisy labels. We will take this up in future. Regarding inference time,  Imponderous Net can  process  at  the  real time rate  of  40 fps in an intel-i7 cpu.

\begin{table}
    \centering 
    \caption{Comparison against SOTA methods for FER} 
    
    \begin{tabular}{|c|c|c|c|c|}
         \hline
          Method      & No. of params   & FERPlus & RAFDB & AffectNet  \\
         \hline
         \hline
         gACNN \cite{gacnn} & 149M & 84.86 & 85.07 & 58.78\\
         RAN \cite{ran} & 11M & 89.16 & 86.9 & 61.71\\
         OADN \cite{oadn} & 24M & 88.71 & 89.83 & 64.06\\
         FMPN \cite{fmpn} & 21.8M & 73.4 & 76.0 & 61.52 \\   
         SCN \cite{scn} & 11M & 88.01 & 87.03 & 64.20\\
         SCAN \cite{scan} & 70M & 89.42 & 89.02 & 65.14\\
         Ours & 1.45M & 88.17 & 87.09 & 62.06\\
         \hline 
    \end{tabular}
    \label{Table 5}
\end{table}

\begin{figure*}
\centerline{\includegraphics[width=5in]{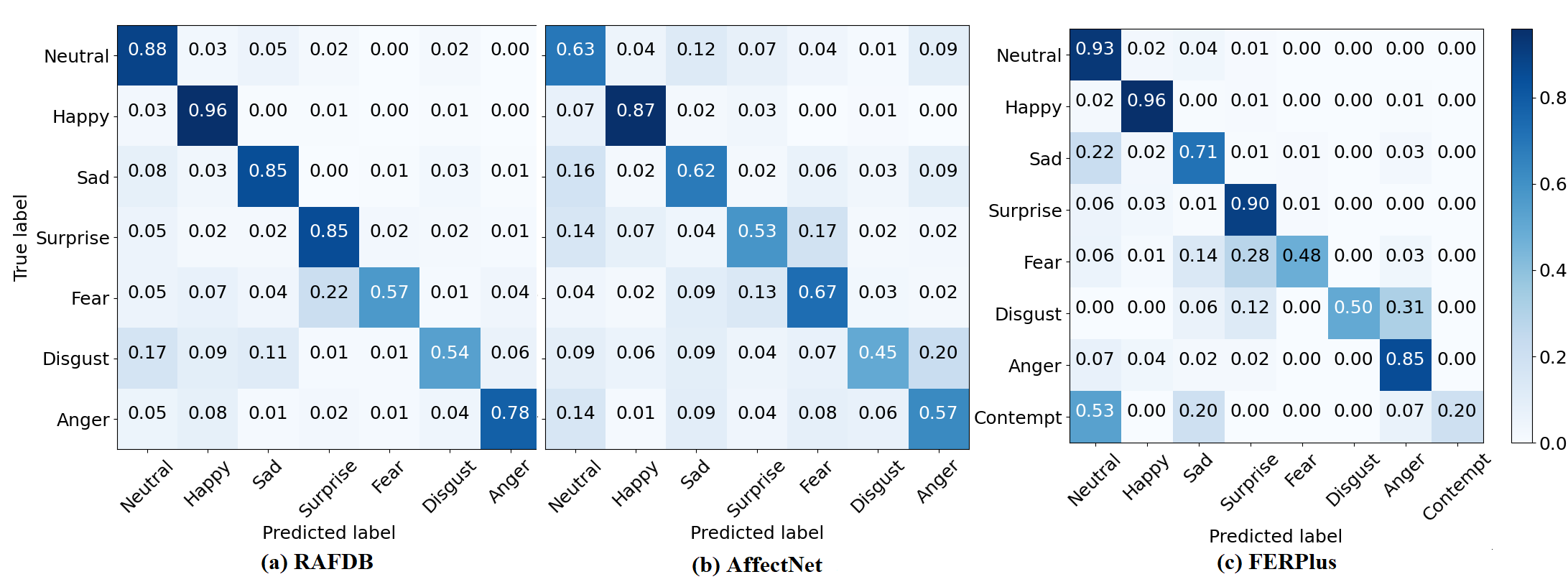}}
\caption{Confusion plots for RAFDB, AffectNet and FERPlus datasets}
\label{confusionplots}
\end{figure*}

\subsection{Performance comparison against light-weight methods}
Table \ref{Table 6} presents the comparison. Note that MobileNet and ShuffleNet are pre-trained on ImageNet. Our Imponderous Net outperforms all the methods on all the datasets. Two important observations can be made from Table \ref{Table 6}. First, too small models perform badly. This is very clear from the performance reported by MicroExpNet and NanoNet. Second, attention mechanism,  local-global context and ensembling does play a significant role under in-the-wild scenario. These components are missing in the light-weight architectures other than ours. We further validate the importance of the second observation in section \ref{ablation-studies}. This emphasizes the importance of our analysis in constructing the Imponderous Net.

\begin{table}
    \centering 
    \caption{Comparison against light-weight methods} 
    
    \begin{tabular}{|c|c|c|c|c|}
         \hline
          Method      & No. of params   & FERPlus & RAFDB & AffectNet  \\
         \hline
         \hline
         MobileNet \cite{mobilenet} & 2.2M & 87.06 & 85.04 & 61.4\\
         ShuffleNet \cite{shufflenet} & 1.3M & 86.77 & 85.3 & 61.1\\
         MicroExpNet \cite{microexpnet} & 65K & 66.82 & 72.36 & 52.74\\
         FaceChannel \cite{facechannel} & 3.5M & 86.71 & 84.42 & 54.62 \\   
         NanoNet \cite{emotionnanonet} & 130K & 80.36 & 72.65 & 59.6\\
         Ours & 1.45M & 88.17 & 87.09 & 62.06\\
         \hline 
    \end{tabular}
    \label{Table 6}
\end{table}

\begin{table*}
    \centering 
    \caption{Performance on challenging subsets of occlusions and pose variations}
    \begin{tabular}{|c|c|c|c|c|c|c|c|c|c|}
         \hline
         \multirow{2}{*}{Method} &  \multicolumn{3}{|c|}{FERPlus} & \multicolumn{3}{|c|}{RAFDB} & \multicolumn{3}{|c|}{AffectNet}  \\ 
         \cline{2-4} \cline{5-7} \cline{8-10}
         
          & Occ. & Pose$>$30 & Pose$>$45 &  Occ. & Pose$>$30 & Pose$>$45 &  Occ. & Pose$>$30 & Pose$>$45 \\
         \hline
         MobileNet & 82.64 & 87.6 & 86.26 & 79.32 & 84.04 & 82.61 & 58.78 & 58.19 &	57.06 \\
         ShuffleNet & 83.8 & 85.64 & 85.64 & 79.59 & 85	& 83.33	& 58.11	& 56.63	& 55.1 \\
         MicroExpNet & 60 &	65.04 &	63.51 &	66.93 &	68.89 & 67.2 & 50.34 & 49.03 &	49.07 \\
         FaceChannel & 80.83 & 84.36 & 81.2 & 75.24 & 81.4 & 78.85 & 58.78 & 56.75 & 55.47 \\
         NanoNet & 74.55 & 79.74 & 77.41 & 63.54 & 71.61 & 68.64 & 55.57 & 56.57 & 56.2 \\
         RAN & 83.63 & 82.23 & 83.63 & 82.72 & 86.74 & 85.2 & 59.12 & 59.05 & 59.37  \\
         OADN & 84.57 & 88.52 & 87.50 & 85.17 & 87.65 & 87.63 & 64.02 & 61.12 & 61.08 \\
         SCAN & 86.12 & 88.89 & 88.15 & 85.03 & 89.82 & 89.07 & 67.06 & 62.64 & 61.31 \\
         Ours & 83.47 & 86.84 & 84.83 & 83.4 & 86.12 & 84.41 & 60.30 & 59.17 & 57.66 \\
         \hline
    \end{tabular}
    \label{Table 7}
\end{table*}

\begin{table}
    \centering 
    \caption{Performance on FED-RO} 
    
    \begin{tabular}{|c|c|}
         \hline
          Method & Performance  \\
         \hline
         \hline
         MobileNet & 60.75 \\
         ShuffleNet &  61\\
         MicroExpNet &  35.25\\
         FaceChannel & 57 \\   
         NanoNet & 56.75 \\
         gACNN & 66.5\\
         RAN & 67.98\\
         OADN & 71.17\\
         SCAN & 73.5\\
         Ours & 65.25 \\
         \hline 
    \end{tabular}
    \label{Table 8}
\end{table}
\subsection{Robustness to occlusions and pose variations}
Table \ref{Table 7} enumerates the performance of our Imponderous Net in comparison to other light-weight methods and SOTA methods for FER under in-the-wild scenario on the challenging subsets of RAFDB, FERPlus and AffectNet with regard to occlusions and pose variations. 

In comparison to light-weight methods, our Imponderous Net has displayed consistent robustness to occlusions and pose variations across all the three datasets. MobileNet and ShuffleNet has done slightly better in a couple of cases in FERPlus. However, they report relatively poor results with regard to occlusions in RAFDB dataset. Lack of consistency in performance of MobileNet and ShuffleNet across datasets  could possibly be due to the absence of the important building blocks we had identified to build Imponderous Net. 

In comparison to SOTA methods for FER under in-the-wild scenario, our Imponderous Net has done on par, or sometimes better than RAN by as large as 4.5\%. It trails the SOTA method SCAN by an average of 3.28\%, except in occlusions subset of AffectNet where the difference is around 6\%. It is to be noted that AffectNet has been a difficult dataset in general, even for high capacity models, since no SOTA method is able to breach even 70\% accuracy. This is because AffectNet possibly has more noisy annotations \cite{scn}.

We also evaluated the performance of our Imponderous Net on real occlusions dataset FED-RO \cite{gacnn}. The results are displayed in Table \ref{Table 8}. Our Imponderous Net outshines all the light-weight methods. Particularly, it has 4.25\% advantage over the next best performing ShuffleNet, again reaffirming the importance of the building blocks in the construction of Imponderous Net. Of course, the performance is relatively far from the current SOTA methods for FER under in-the-wild setting like OADN and SCAN. Overall, performance comparison against SOTA methods for FER under in-the-wild setting in Table \ref{Table 7} and Table \ref{Table 8} indicate that challenging conditions does require a relatively larger number of parameters for performance boost. Nevertheless, our Imponderous Net has exhibited consistent superior robustness over all the light-weight methods under challenging conditions. It is easily extensible (see Fig. \ref{framework}), and hence can be experimented with extensions, for further improvement under challenging conditions. This will be a part of the future work.

\subsection{Expression discrimination}
Confusion matrices of Imponderous Net on all the three datasets are shown in Fig. \ref{confusionplots}. Happiness is the easiest recognizable expression on all the datasets. Surprise is relatively easily recognizable in RAFDB and FERPlus. Fear is relatively easily recognizable in AffectNet. While disgust is the most difficult expression to recognize in RAFDB and AffectNet datasets, contempt pulls down the performance in FERPlus dataset. Disgust is generally confused with anger in AffectNet and FERPlus; it is confused with neutral in RAFDB. 

\subsection{Other ablation studies}
\label{ablation-studies}
\begin{figure}
\centerline{\includegraphics[width=2in]{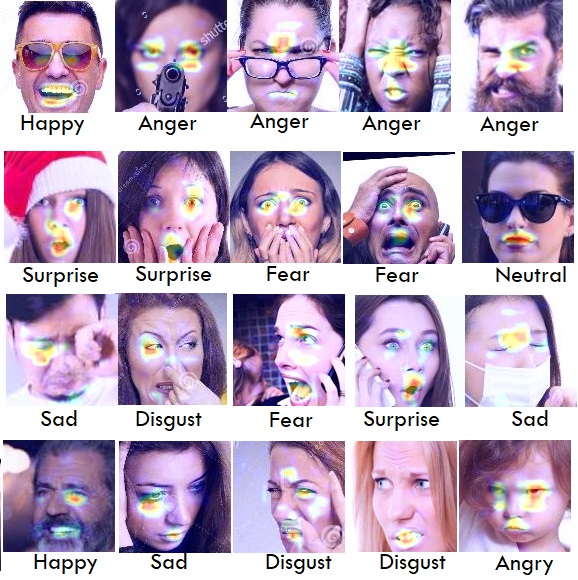}}
\caption{Visualization of activation maps using gradcam \cite{Selvaraju_60} (Red color indicates high activation and blue color indicates low activation).}
\label{gradcam}
\end{figure}

We analyse the influence of ECA, implicit ensembling, global context and the placement of ECA in Imponderous Net. By placement of ECA, we mean whether ECA is placed after region partitioning or before region partitioning.  We investigate all these on RAFDB dataset. Table \ref{Table 9} presents the results. Presence of ECA enhances the performance by around 0.7\%. Even though this gain might look small in the absolute sense, it is significant in the relative sense in narrowing the gap with performance of higher capacity SOTA methods for FER under in-the-wild setting and widening the gap with performance of other light-weight methods (see column 4 in Tables \ref{Table 5} and \ref{Table 6}). Without implicit ensembling, there is almost a 3\% reduction in performance. Without incorporating global context, performance diminishes by 1.07\%. By placing ECA prior to region partitioning, performance downgrades by 1.2\%. Note that, whether ECA is done before region partioning or as in Fig. \ref{framework}, the number of parameters introduced by ECA will remain the same. However, ECA placed prior to partitioning will do a global average pool of the entire 16 x 16 spatial output. By this, we lose on local attention. Further partitioning does not add value in this scenario since each partition has a corrupted local context due to weighting by global attention weights. In fact, not having ECA is better than this scenario. We also visualize the activation maps using gradcam \cite{Selvaraju_60}, which are shown in Fig.\ref{gradcam}. It is clear that Imponderous Net has avoided occlusions, and also handled pose variations very well.

\begin{table}
    \centering 
    \caption{Other ablation studies} 
    
    \begin{tabular}{|c|c|}
         \hline
          Method & Performance \\
          \hline
         Without ECA & 86.38 \\
         Without implicit ensembling & 84.29 \\
         Without global context & 86.02 \\
         ECA before partition & 85.89 \\
         Ours & 87.09\\
         \hline
         
    \end{tabular}
    \label{Table 9}
\end{table}

\section{Conclusions}
Imponderous Net has been carefully built using important components including light-weight effective base network, attention unit, local-global context units and implicit ensembling. Overall, it has around 1.45M parameters only. We have demonstrated its power packed performance through extensive comparison against light-weight and heavy SOTA methods. We also pointed out that challenging conditions demand relatively larger number of parameters. We believe that the extensive comparison results we have illustrated would be very useful to the FER research community.

\bibliographystyle{unsrt}  
\bibliography{references}  %%% Remove comment to use the external .bib file (using bibtex).

\begin{thebibliography}{10}

\bibitem{56}
Jaffar MA.
\newblock Facial expression recognition using hybrid texture features based
  ensemble classifier.
\newblock {\em International Journal of Advanced Computer Science and
  Applications(IJACSA)}, 6, 2017.

\bibitem{60}
Hok chun Lo and R.~Chung.
\newblock Facial expression recognition approach for performance animation.
\newblock {\em IEEE Proceedings Second International Workshop on Digital and
  Computational Video}, 6:613--622, Feb 2001.

\bibitem{63_a}
T.~Kanade, J.~F. Cohn, and Y.~Tian.
\newblock Comprehensive database for facial expression analysis.
\newblock In {\em Proceedings of the Foruth IEEE International Conference on
  Automatic Face and Gesture Recognition(FG'OO)}, pages 46--53, Grenoble,
  France, 2000.

\bibitem{63_b}
P.~Lucey, J.~F. Cohn, T.~Kanade, J.~Saragih, Z.~Ambadar, and I.~Matthews.
\newblock The extended cohn-kanade dataset (ck+): A complete dataset for action
  unit and emotion-specified expression.
\newblock In {\em Computer Vision and Pattern Recognition Workshops(CVPRW)},
  page 94–101, San Francisco, USA, 2010.

\bibitem{61}
G.~Zhao, X.~Huang, M.~Taini, S.~Z. Li, and M.~Pietikainen.
\newblock Facial expression recognition from near-infrared videos.
\newblock {\em Image and Vision Computing}, 29:607--619, 2011.

\bibitem{58}
Shih FY, Chuang CF, and Wang PSP.
\newblock Performance comparisons of facial expression recognition in jaffe
  database.
\newblock {\em International Journal of Pattern Recognition and Artificial
  Intelligence}, 22:445--459, 2008.

\bibitem{40}
H.~Yang, U.~Ciftci, and L.~Yin.
\newblock Facial expression recognition by de-expression residue learning.
\newblock {\em CVPR}, page 2168–2177, 2018.

\bibitem{chen2019facial_51}
Yuedong Chen, Jianfeng Wang, Shikai Chen, Zhongchao Shi, and Jianfei Cai.
\newblock Facial motion prior networks for facial expression recognition.
\newblock {\em IEEE Visual Communications and Image Processing (VCIP)}, 2019.

\bibitem{scan}
Darshan Gera and S~Balasubramanian.
\newblock Landmark guidance independent spatio-channel attention and
  complementary context information based facial expression recognition.
\newblock {\em Pattern Recognition Letters}, 145:58--66, 2021.

\bibitem{oadn}
Hui Ding, Peng Zhou, and Rama Chellappa.
\newblock Occlusion-adaptive deep network for robust facial expression
  recognition.
\newblock In {\em 2020 IEEE International Joint Conference on Biometrics
  (IJCB)}, pages 1--9. IEEE, 2020.

\bibitem{ran}
Kai Wang, Xiaojiang Peng, Jianfei Yang, Debin Meng, and Yu~Qiao.
\newblock Region attention networks for pose and occlusion robust facial
  expression recognition.
\newblock {\em IEEE Transactions on Image Processing}, 29:4057-- 4069, January
  2020.

\bibitem{gacnn}
Yong Li, Jiabei Zeng, Shiguang Shan, and Xilin Chen.
\newblock Occlusion aware facial expression recognition using cnn with
  attention mechanism.
\newblock {\em IEEE Transactions on Image Processing}, 28:2439--2450, May 2019.

\bibitem{autistic}
A.~{Sarrafzadeh}, J.~{Shanbehzadeh}, F.~{Dadgostar}, C.~{Fan}, and
  S.~{Alexander}.
\newblock Assisting the autistic with real-time facial expression recognition.
\newblock In {\em 2009 International Conference on Innovations in Information
  Technology (IIT)}, pages 90--94, 2009.

\bibitem{rafdba}
Shan Li and Weihong Deng.
\newblock Reliable crowdsourcing and deep locality-preserving learning for
  unconstrained facial expression recognition.
\newblock {\em IEEE Transactions on Image Processing}, 28(1):356--370, 2019.

\bibitem{rafdbb}
Shan Li, Weihong Deng, and JunPing Du.
\newblock Reliable crowdsourcing and deep locality-preserving learning for
  expression recognition in the wild.
\newblock In {\em 2017 IEEE Conference on Computer Vision and Pattern
  Recognition (CVPR)}, pages 2584--2593. IEEE, 2017.

\bibitem{ferplus}
E.~Barsoum, C.~Zhang, C.~C. Ferrer, and Z.~Zhang.
\newblock Training deep networks for facial expression recognition with
  crowdsourced label distribution.
\newblock {\em In Proceedings of the 18th ACM International Conference on
  Multimodal Interaction}, page 279–283, 2016.

\bibitem{affectnet}
Ali Mollahosseini, Behzad Hasani, and Mohammad~H Mahoor.
\newblock Affectnet: A database for facial expression, valence, and arousal
  computing in the wild.
\newblock {\em IEEE Transactions on Affective Computing}, 2017.

\bibitem{microexpnet}
Ilke Cugu, Eren Sener, and Emre Akbas.
\newblock Microexpnet: An extremely small and fast model for expression
  recognition from face images.
\newblock In {\em 2019 Ninth International Conference on Image Processing
  Theory, Tools and Applications (IPTA)}, pages 1--6. IEEE, 2019.

\bibitem{facechannel}
Pablo Barros, Nikhil Churamani, and Alessandra Sciutti.
\newblock The facechannel: A fast and furious deep neural network for facial
  expression recognition.
\newblock {\em SN Computer Science}, 1(6):1--10, 2020.

\bibitem{emotionnanonet}
James Ren~Hou Lee, Linda Wang, and Alexander Wong.
\newblock Emotionnet nano: An efficient deep convolutional neural network
  design for real-time facial expression recognition.
\newblock {\em arXiv preprint arXiv:2006.15759}, 2020.

\bibitem{imagenet_cvpr09_50}
J.~Deng, W.~Dong, R.~Socher, L.-J. Li, K.~Li, and L.~Fei-Fei.
\newblock {ImageNet: A Large-Scale Hierarchical Image Database}.
\newblock In {\em CVPR09}, 2009.

\bibitem{17}
Yandong Guo, Lei Zhang, Yuxiao Hu, Xiaodong He, and Jianfeng Gao.
\newblock Ms-celeb-1m: A dataset and benchmark for large-scale face
  recognition.
\newblock {\em ECCV}, 2016.

\bibitem{20}
Q.~Cao, L.~Shen, W.~Xie, O.~M. Parkhi, and A.~Zisserman.
\newblock Vggface2: A dataset for recognising face across pose and age.
\newblock {\em International Conference on Automatic Face and Gesture
  Recognition}, 2018.

\bibitem{cnn_features}
Jun-Cheng Chen, Vishal~M Patel, and Rama Chellappa.
\newblock Unconstrained face verification using deep cnn features.
\newblock In {\em 2016 IEEE winter conference on applications of computer
  vision (WACV)}, pages 1--9. IEEE, 2016.

\bibitem{lightcnn}
Xiang Wu, Ran He, Zhenan Sun, and Tieniu Tan.
\newblock A light cnn for deep face representation with noisy labels.
\newblock {\em IEEE Transactions on Information Forensics and Security},
  13(11):2884--2896, 2018.

\bibitem{lightcnn_ihbition}
S~Amari.
\newblock Dynamics of pattern formation in lateral-inhibition type neural
  fields.
\newblock {\em Biological cybernetics}, 27(2):77--87, 1977.

\bibitem{dn_memorize}
Devansh Arpit, Stanis{\l}aw Jastrz{\k{e}}bski, Nicolas Ballas, David Krueger,
  Emmanuel Bengio, Maxinder~S Kanwal, Tegan Maharaj, Asja Fischer, Aaron
  Courville, Yoshua Bengio, et~al.
\newblock A closer look at memorization in deep networks.
\newblock In {\em International Conference on Machine Learning}, pages
  233--242. PMLR, 2017.

\bibitem{eca}
Qilong Wang, Banggu Wu, Pengfei Zhu, Peihua Li, Wangmeng Zuo, and Qinghua Hu.
\newblock Eca-net: Efficient channel attention for deep convolutional neural
  networks.
\newblock {\em CVPR}, 2019.

\bibitem{sqe}
Jie Hu, Li~Shen, and Gang Sun.
\newblock Squeeze-and-excitation networks.
\newblock {\em CVPR}, pages 7132--7141, 2018.

\bibitem{Hassner1}
Tal~Hassner Gil~Levi.
\newblock Emotion recognition in the wild via convolutional neural networks and
  mapped binary patterns.
\newblock {\em International Conference on Multimodal Interaction}, pages
  503--510, 2015.

\bibitem{mtcnn}
K.~Zhang, Z.~Zhang, Z.~Li, and Y.~Qiao.
\newblock Joint face detection and alignment using multitask cascaded
  convolutional networks.
\newblock {\em IEEE Signal Processing Letters}, 23(10):1499--1503, 2016.

\bibitem{casia}
Dong Yi, Zhen Lei, Shengcai Liao, and Stan~Z Li.
\newblock Learning face representation from scratch.
\newblock {\em arXiv preprint arXiv:1411.7923}, 2014.

\bibitem{autoaugument}
Ekin~Dogus Cubuk, Barret Zoph, Dandelion Mane, Vijay Vasudevan, and Quoc~V. Le.
\newblock Autoaugment: Learning augmentation policies from data.
\newblock {\em CVPR}, pages 113--123, 2019.

\bibitem{scn}
Kai Wang, Xiaojiang Peng, Jianfei Yang, Shijian Lu, and Yu~Qiao.
\newblock Suppressing uncertainties for large-scale facial expression
  recognition.
\newblock In {\em CVPR}, pages 6897--6906, 2020.

\bibitem{fmpn}
Yuedong Chen, Jianfeng Wang, Shikai Chen, Zhongchao Shi, and Jianfei Cai.
\newblock Facial motion prior networks for facial expression recognition.
\newblock {\em VCIP}, 2019.

\bibitem{mobilenet}
Mark Sandler, Andrew Howard, Menglong Zhu, Andrey Zhmoginov, and Liang-Chieh
  Chen.
\newblock Mobilenetv2: Inverted residuals and linear bottlenecks.
\newblock {\em CVPR}, pages 4510--4520, 2018.

\bibitem{shufflenet}
Xiangyu Zhang, Xinyu Zhou, Mengxiao Lin, and Jian Sun.
\newblock Shufflenet: An extremely efficient convolutional neural network for
  mobile devices.
\newblock {\em CVPR}, pages 6848--6856, 2018.

\bibitem{Selvaraju_60}
Ramprasaath~R Selvaraju, Michael Cogswell, Abhishek Das, Ramakrishna Vedantam,
  Devi Parikh, and Dhruv Batra.
\newblock Grad-cam: Visual explanations from deep networks via gradient-based
  localization.
\newblock In {\em Proceedings of the IEEE international conference on computer
  vision}, pages 618--626, 2017.

\end{thebibliography}

\end{document}